# Detection of DDoS Attacks in Software Defined Networking Using Machine Learning Models

Ahmad Hamarshe, Huthaifa I. Ashqar*, Mohammad Hamarsheh
Department of Natural-Engineering and Technology Sciences, Arab American University, Palestine
a.hamarshe1@student.aaup.edu, Huthaifa.ashqar@aaup.edu, Mohammad.hamarsheh@aaup.edu

Abstract

The concept of Software Defined Networking (SDN) represents a modern approach to networking that separates the control plane from the data plane through network abstraction, resulting in a flexible, programmable and dynamic architecture compared to traditional networks. The separation of control and data planes has led to a high degree of network resilience, but has also given rise to new security risks, including the threat of distributed denial-of-service (DDoS) attacks, which pose a new challenge in the SDN environment.

In this paper, the effectiveness of using machine learning algorithms to detect distributed denial-of-service (DDoS) attacks in software-defined networking (SDN) environments is investigated. Four algorithms, including Random Forest, Decision Tree, Support Vector Machine, and XGBoost, were tested on the CICDDoS2019 dataset, with the timestamp feature dropped among others. Performance was assessed by measures of accuracy, recall, accuracy, and F1 score, with the Random Forest algorithm having the highest accuracy, at 68.9%. The results indicate that ML-based detection is a more accurate and effective method for identifying DDoS attacks in SDN, despite the computational requirements of non-parametric algorithms.

## I  Introduction

Distributed Denial of Service (DDOS) attacks are a common and costly form of cyber-attacks that aim to disrupt the availability of a network or service by overwhelming it with traffic from multiple sources [1]. DDOS attacks can be launched using various tactics and techniques, such as flooding the network with packets or requests, or exploiting vulnerabilities in the network protocols or devices [2]. DDOS attacks can cause significant damage and disruption to businesses, organizations, and individuals, and can affect the reputation and trust of the targeted entities [3].

Software-Defined Networking (SDN) is a promising paradigm that enables the centralized control and management of network resources and has the potential to enhance the security and resilience of networks against DDOS attacks [3]. In SDN, the control plane and data plane are separated, and the control plane is implemented in software, which enables the dynamic and flexible configuration and management of the network [4]. However, SDN itself may be vulnerable to DDOS attacks, and there is a need for effective and efficient methods to detect and mitigate DDOS attacks in SDN [5]. To mitigate DDOS attacks, various detection and defense mechanisms have been proposed, including traditional intrusion detection systems (IDS) and firewalls, as well as more advanced techniques such as machine learning (ML) [6].

Machine learning algorithms are a powerful tool for detecting and classifying network traffic and have been widely applied to DDOS attack detection in SDN [6]. The efficiency of machine learning algorithms for DDOS attack detection in SDN depends on the specific algorithms, datasets, and evaluation methods that are used [7]. Different machine learning algorithms have different strengths and limitations and may perform differently depending on the characteristics and quality of the data [8]. In this paper, four machine learning algorithms (Random Forest, SVM, Decision Tree, and XGBoost) will be applied due to their ability to learn from and classify large amounts of data. However, the relative effectiveness of these algorithms has not yet been comprehensively compared in the context of DDOS attack detection in SDN. In addition, most of the previous studies did not rule out some features that are not feasible when applied to live network traffic. To address this gap in the literature, the current study aims to compare the performance of machine learning algorithms for detecting DDOS attacks before they occur in SDN and to consider the selection of appropriate features to obtain true accuracy for each. This comparison will be made using a dataset of simulated DDOS attacks and normal network traffic and will take into account a range of performance metrics including accuracy, precision, recall and f1-score.

## II Background and Related Works

### A DDOS Attacks and their Detection

In this section, a brief overview of Distributed Denial of Service (DDoS) attacks is presented, along with a discussion on some of the existing and related work on DDoS attack detection. DDoS attacks are a form of cyber-attack that aims to reduce service availability by overwhelming a network's resources, making it difficult or impossible for authorized users to access their services. These attacks can be classified into two types: reflection-based and exploitation-based. Reflection-based attacks use an authorized third-party component to conceal the attacker's identity and overload the victim with response packets. Exploitation-based attacks, on the other hand, are conducted by the component of the third party [9].

In terms of detection methods, DDoS attacks can be categorized into three types: Signature-Based (SB), Anomaly-Based (AB), and entropy-based (EB) approaches. SB frameworks rely on measuring attributes across a signature database or from recognized malicious threats, making them easy to implement but less effective against evolving threats. AB systems use data mining and statistical approaches to track and compare network traffic against a defined baseline but may generate false positive alerts if the baseline is not properly set up. EB systems, on the other hand, rely on changes in entropy values to detect anomalous incidents, making them more sensitive than the other two types. However, in high-speed networks, an effective algorithm is required to reduce processing time and memory consumption.

### B Related Work

In recent years, the use of machine learning (ML) techniques to detect and mitigate distributed denial-of-service (DDoS) attacks has gained significant attention in the field of computer science. A number of studies have proposed various machine learning-based approaches to identifying DDoS attacks, utilizing different datasets and performance metrics. One of the most prominent studies in this area [10] a study on the effectiveness of using 25 time-based features to detect and classify 12 types of DDoS attacks using machine learning classifiers. The study found that the majority of models achieved an accuracy of around 99% in detecting DDoS attacks and an accuracy of around 70% in classifying specific DDoS attack types. Furthermore, the study found that using a smaller subset of time-based features, which reduces training time, still achieved high accuracy in detecting DDoS attacks. Another notable study [11] employed a machine-learning algorithm to detect DDoS attacks in an SDN environment. The authors reported similar results, with accuracy rates above 99%, and demonstrated that the Decision Tree (DT) algorithm had the highest accuracy rate of 99.90%. [12] proposed a new scheme for DDoS data collection in SDN by collecting a largely static data model using port statistics. The study found that the Support Vector Machine (SVM) with linear and radial basis function kernels was the most effective way to successfully identify a DDoS attack, with an accuracy rate of close to 100%. [13] compared the performance of different ML algorithms for detecting DDoS attacks and found that the K-Nearest Neighbor (KNN) and Decision Tree (DT) algorithms had high accuracy in detecting DDoS attacks. However, a comparative analysis showed that DT had a higher accuracy of 99.91% compared to KNN's 98.94% and also had a better running time. [14] suggested using ML-based approaches to detect DDoS attacks in SDN and utilized the KDD99 dataset to train and test the model. The study found that the decision tree algorithm and support vector machine (SVM) provided an 85% better accuracy and detection rate. [15] proposed using the ensemble algorithm to detect a DDoS attack in an SDN environment. The study used in SDN as the dataset and used K-means++ and Random Forest to obtain high detection accuracy and efficiency. The proposed method achieved 100% value accuracy, 100% accuracy, 100% retrieval, and 100% F1 scale with a processing time of 1.5 s. [16] proposed the use of ML-based methods to detect DDoS attacks in wireless networks. The study used the decision tree and the KDDCup'99 dataset for classification, and results showed that the J48 algorithm had a high accuracy of 99.94% in detecting DDoS attacks.

However, in [17] the authors developed architecture for identifying and a mitigating LR-DDoS attack in SDN, which includes IDS trained using machine learning models and has a high detection rate (95%). The architecture was tested in a simulated environment designed to closely mimic real-world production networks. The IDS was able to successfully mitigate all detected attacks. Also, in [18], the authors proposed DNN-based IDS for real-time detection of DDoS attacks in SDN, which was efficient and achieves a high detection accuracy of 97.59%. The IDS was designed to counter the use of sophisticated techniques by attackers to launch DDoS attacks on vulnerable SDN networks While these studies have made great strides in using ML algorithms for DDoS detection in SDN, there is still room for improvement. Many previous studies focused on a limited number of features, some of which used features that are useless when using the model in real time, in addition to being limited to a limited number of machine learning algorithms and did not compare the performance of these algorithms (Random Forest, SVM, Decision Tree, and XGBoost) way enough between each other before in the CIC-DDOS2019 dataset. To address these limitations, this paper aims to generate DDoS detection identifiers in SDN using these ML algorithms considering the use of the most effective real-time features and to comprehensively evaluate the performance of these algorithms in terms of accuracy and efficiency.

## III  Dataset and ML Model

### A  Dataset

The CIC-DDOS2019 dataset is a comprehensive dataset that was created by the Canadian Institute of Cybersecurity (CIC) to support research on detecting and mitigating Distributed Denial of Service (DDoS) attacks. The dataset contains a wide variety of DDoS attack types, including TCP, UDP, and HTTP floods, as well as botnet-based attacks. The dataset also includes both benign and malicious traffic, making it ideal for training and evaluating machine learning-based DDoS detection systems. Additionally, the dataset is large and diverse, with over 2 million flow records and over 3 million packets, providing a significant amount of data for research. The dataset includes detailed information about the attack parameters, such as the type of attack, the source and destination IP addresses, and the packet size. This information can be used to gain a deeper understanding of the characteristics of different DDoS attack types and to develop more effective detection and mitigation strategies [9].

### B  Machine Learning Model

1) Support Vector Machine (SVM) is a machine learning algorithm designed for use in both classification and regression problems. Its primary objective is to find a hyperplane in an n-dimensional feature space that accurately separates the data into distinct classes. The learning process of SVM involves two phases. Firstly, the input data is mapped into an n-dimensional space where each feature is treated as a support vector. Secondly, the best hyperplane for class separation is determined through optimization. To handle complex non-linear relationships, SVM makes use of the "Kernel trick," which simplifies the computational process [19].

2) Random Forest (RF) is a type of ensemble learning algorithm in supervised machine learning. It uses a large number of decision trees to produce stable and precise predictions. RF creates a "forest" of decision trees, typically trained using the "bagging" method. The idea behind bagging is to combine multiple learning models to enhance the overall outcome [20].

3) Decision Tree (DT) is a straightforward approach in machine learning that's used for supervised learning problems. The algorithm builds a tree-like structure to map out the relationships between different inputs and the outputs they lead to. The input data is separated into branches based on its values, and the leaf nodes of the tree indicate the final prediction for a class or numerical value. One of the great things about Decision Trees is that they're easy to understand and interpret, which makes them a valuable tool for figuring out which inputs are important and how they influence the final predictions [21].

4) XGBoost is an advanced machine learning technique that's used in supervised learning. It's based on a concept called Gradient Boosting, which means it takes multiple decision trees and combines their predictions to produce even more accurate results. XGBoost is a powerful tool that can handle large amounts of data and complex relationships between inputs and outputs with ease. It's become very popular in data science competitions and is widely used in many real-world applications [22].

## IV  Proposed Approach

Dealing with Distributed Denial of Service (DDoS) attacks is a top priority when it comes to software-defined networking (SDN) due to its centralized controller architecture. The constant evolution of these attacks requires updated and innovative systems to respond effectively, and this is where the evaluation of machine learning (ML) methods for DDoS detection comes into play. By training the detection system to learn traffic patterns from new information, ML offers a more accurate and efficient solution compared to other detection methods. There are three main categories for detecting DDoS attacks, based on the detection metric and mechanism used: Information-theory based detection, ML-based detection, and ANN-based detection. For this study, we opted for ML-based detection due to its ease of implementation and relatively high degree of precision compared to Information-theory based models. The standard CICDDoS2019 dataset was used to train the ML models, and the network data was fed to the trained model to predict whether the data was anomalous or benign. This concept is depicted in Figure 1.

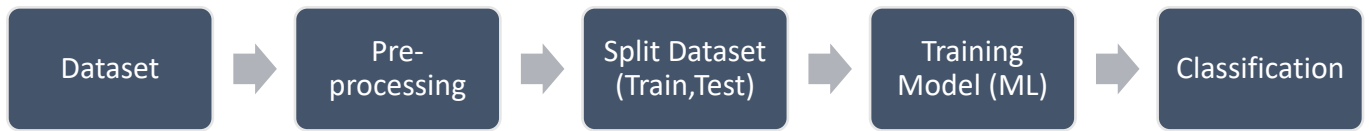

Fig. 1. Proposed Methodology

### A  Machine Learning Approach

Multiple machine learning (ML) techniques are employed to detect DDoS attacks. Four classification algorithms, including Random Forest (RF) classifier, Decision Tree (DT), Support Vector Machine (SVM), and XGBoost (GBDT), are selected based on the criteria of non-parametric algorithms. Non-parametric algorithms are a powerful tool in the world of machine learning, allowing us to capture the nuances of complex relationships between input and output variables [23], [24]. These algorithms are known for their flexibility and ability to handle a wide range of data types. However, this flexibility doesn't come without its challenges. When dealing with large datasets, non-parametric algorithms can be quite demanding computationally. To help manage this, we often employ techniques such as pruning or adding a penalty term to the objective function. Despite these efforts, non-parametric algorithms are still slower than their parametric counterparts, like linear regression, which can make them less suitable for real-time, large-scale applications. Despite these limitations, non-parametric algorithms remain a valuable tool in our machine learning toolbox. By offering the ability to model complex relationships, these algorithms have the potential to provide deep insights into a variety of problems. With ongoing efforts to optimize computational efficiency, we can only expect non-parametric algorithms to continue making a big impact in the world of machine learning.

### B  CIC-DDOS2019 Dataset for Training Model

The CICDDoS2019 dataset [20] was used in this research to train and evaluate the proposed model. This dataset contains both benign and up-to-date common DDoS attacks, providing a true-to-life representation of network traffic. The data is analyzed using Pandas (python library), because it provides data structures for efficiently storing large datasets and tools for working with them. In this research, Pandas was used to analyze the CICDDoS2019 dataset, which contains network traffic data with multiple variables. The library's powerful data manipulation capabilities, such as indexing, grouping, and merging, allowed for effective analysis and processing of the complex dataset. after pre-processing to dataset, it labeled based on 20 features, (destination IP, flow ID, minimum forward segment size, flow duration, forward header length, source port, minimum packet length, minimum forward packet length, mean packet length, maximum forward packet length, average packet size, average forward segment size, mean forward packet length, maximum packet length, total forward inter-arrival time, total length of forward packets, sub flow forward bytes, destination port, and source IP) [9].

However, we chose to exclude certain features, such as the timestamp, which was frequently used in prior studies [timestamp]. Despite being one of the most important features during feature extraction, we aimed to demonstrate the realism of our algorithms by not relying solely on the timestamp. Previous research has shown that algorithms relying on timestamp have a high accuracy in detecting the type of attack, near 99%, however, this approach is not practical in real-world scenarios as timestamps in actual network traffic will differ, and thus this model was not fit and did not provide high performance.

Various attack types were considered with the exception of WebDDoS due to insufficient data (less than 400 records). The considered attacks included DrDoS_DNS, DrDoS_LDAP, DrDoS_MSSQL, DrDoS_NetBIOS, DrDoS_NTP, Portmap, DrDoS_SNSSDP, TFTP, DrDoS_UDP and UDP-lag, in addition to BENIGN records. An equal number of Benign and malignant records, 113,828 each, were utilized to achieve the highest accuracy in modeling, resulting in a total of 1,365,936 records. The data was split into 80% for training the model and 20% for testing [9].

### C  Performance Measurements

The accuracy of classification algorithms was assessed using four metrics, namely accuracy, precision, recall, and F1 score[25]. These evaluation scales are based on four key elements: True Positives (TP), which represents the number of DDoS traffic instances correctly classified; True Negatives (TN), referring to the number of normal traffic instances accurately classified; False Positives (FP), indicating the number of DDoS traffic instances misclassified; and False Negatives (FN), denoting the number of normal traffic instances misclassified. The efficiency of the algorithms was calculated using a confusion matrix, as depicted in Table 1.

1) The Confusion Matrix reflects the accuracy of classifying generated data into pre-determined categories throughout the learning, validation, and testing process.

2) Accuracy represents the proportion of DDoS attacks accurately identified in the dataset and is calculated according to Eq. (1).

$$\text{Accuracy} = \frac{P+Q}{P+Q+R+S} \quad (1)$$

3) Recall quantifies the ability to correctly classify DDoS traffic and is calculated according to Eq. (2).

$$\text{Recall} = \frac{P}{P+S} \quad (2)$$

4) Precision reflects the proportion of correctly categorized DDoS attacks out of all instances classified as DDoS and is determined by Eq. (3).

$$\text{Precision} = \frac{P}{P+R} \quad (3)$$

5) F1-score is a composite measure of performance that balances Precision (A) and Recall (B) through a harmonic mean, as specified by Eq. (4).

$$\text{Precision} = 2 * \frac{A*B}{A+B} \quad (4)$$

*Table 1. Confusion Matrix*

|  |  | Predicted | |
|---|---|---|---|
|  |  | Positive | Negative |
| *Actual* | Positive | P | S |
|  | Negative | R | Q |

## V  Result and Analysis

The performance of the machine learning (ML) model applied to the CICDDOS2019 dataset was evaluated utilizing a suite of metrics, including F1 score, accuracy, recall, and precision. These metrics offer a comprehensive evaluation of the model's capability to detect various types of DDoS attacks with accuracy. The accuracy results, ranked in descending order, revealed that the Random Forest algorithm had the highest accuracy of 68.9%, followed by XGBoost with 63.5%, Support Vector Machine with 59.2%, and Decision Tree with 47.8%. As such, Random Forest emerged as the most accurate among the evaluated algorithms.

The results of the Random Forest algorithm are displayed in Table (2), revealing its low detection accuracy for some attack types, such as DrDoS_SSDP with an accuracy of 63% and DrDoS_UDP with an accuracy of 49%. On the other hand, the model achieved high accuracy for certain attack types, such as Portmap, with an accuracy of 99%. Table (3) presents the results obtained by the Decision Tree algorithm, which yielded lower accuracy scores compared to the Random Forest algorithm. Despite this, the Decision Tree model displayed a high accuracy in detecting benign records, reaching over 96% accuracy. Similarly, the model exhibited a high accuracy of 98% for the Portmap attack, but failed to accurately detect certain attacks, such as DrDoS_SNMP and DrDoS_SSDP, with accuracy scores below 30%.

*Table 2. Classification Report for Random Forest*

| Label | Precision | Recall | F1-score |
|---|---|---|---|
| BENIGN | 0.98 | 1.00 | 0.99 |
| DrDoS_DNS | 0.55 | 0.62 | 0.58 |
| DrDoS_LDAP | 0.56 | 0.78 | 0.65 |
| DrDoS_MSSQL | 0.70 | 0.77 | 0.73 |
| DrDoS_NTP | 0.90 | 0.90 | 0.90 |
| DrDoS_NetBIOS | 0.82 | 0.21 | 0.33 |
| DrDoS_SNMP | 0.66 | 0.94 | 0.78 |
| DrDoS_SSDP | 0.63 | 0.13 | 0.22 |
| DrDoS_UDP | 0.49 | 0.92 | 0.64 |
| Portmap | 0.99 | 1.00 | 1.00 |
| Syn | 0.56 | 0.80 | 0.66 |
| TFTP | 0.70 | 0.54 | 0.61 |
| UDP-lag | 0.94 | 0.35 | 0.51 |

Table 3. Classification Report for Decision Tree

| Label | Precision | Recall | F1-score |
|---|---|---|---|
| BENIGN | 0.96 | 0.99 | 0.98 |
| DrDoS_DNS | 0.11 | 0.33 | 0.16 |
| DrDoS_LDAP | 0.44 | 0.56 | 0.49 |
| DrDoS_MSSQL | 0.42 | 0.46 | 0.44 |
| DrDoS_NTP | 0.82 | 0.34 | 0.48 |
| DrDoS_NetBIOS | 0.39 | 0.21 | 0.28 |
| DrDoS_SNMP | 0.31 | 0.02 | 0.03 |
| DrDoS_SSDP | 0.27 | 0.08 | 0.12 |
| DrDoS_UDP | 0.40 | 0.76 | 0.53 |
| Portmap | 0.98 | 0.98 | 0.98 |
| Syn | 0.55 | 0.70 | 0.62 |
| TFTP | 0.63 | 0.44 | 0.52 |
| UDP-lag | 0.79 | 0.37 | 0.51 |

Moreover, the Receiver Operating Characteristic (ROC) curve was employed to visually analyze the balance between the true positive rate and false positive rate of the model, thereby illuminating its capability in differentiating between benign and malicious traffic. Furthermore, the accuracy, recall, precision, and F1 score were calculated and reported for each label in the dataset, offering an in-depth analysis of the model's performance for each type of attack. Such information is imperative in evaluating the overall performance of the model and recognizing areas for enhancement. Figure 1 presents the ROC curve of the highest-performing Random Forest model, while Figure 2 displays the ROC curve of the lowest-performing Decision Tree model.

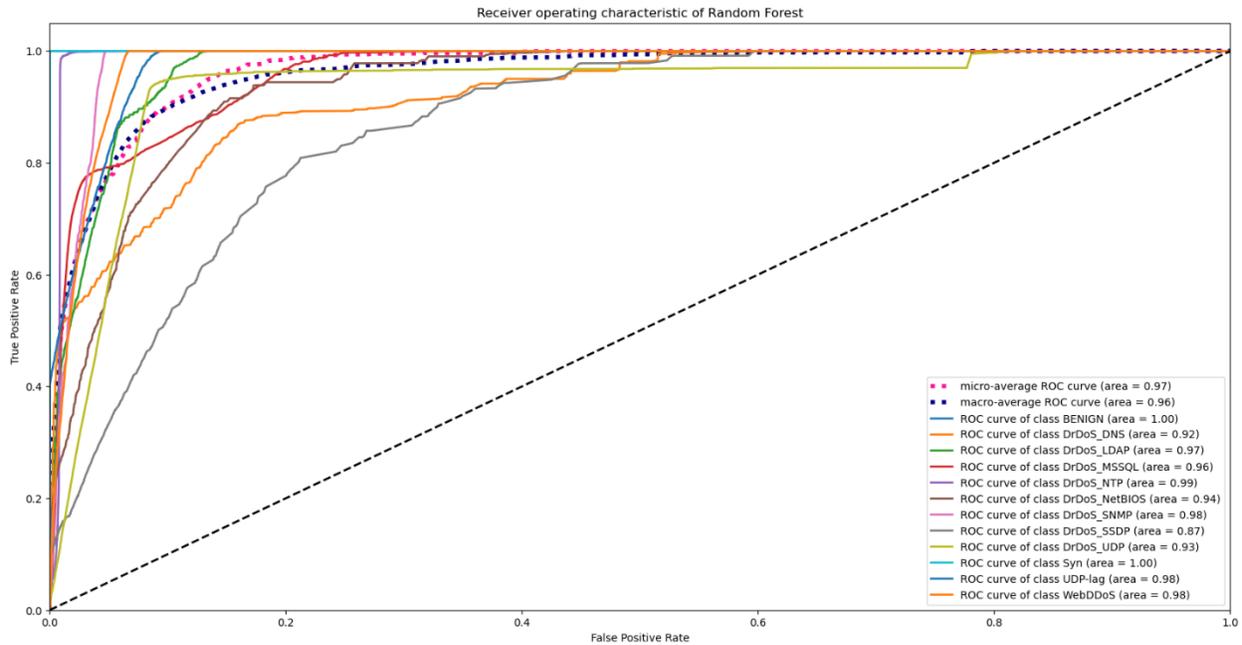

Fig. 2. ROC Curve of Random Forest

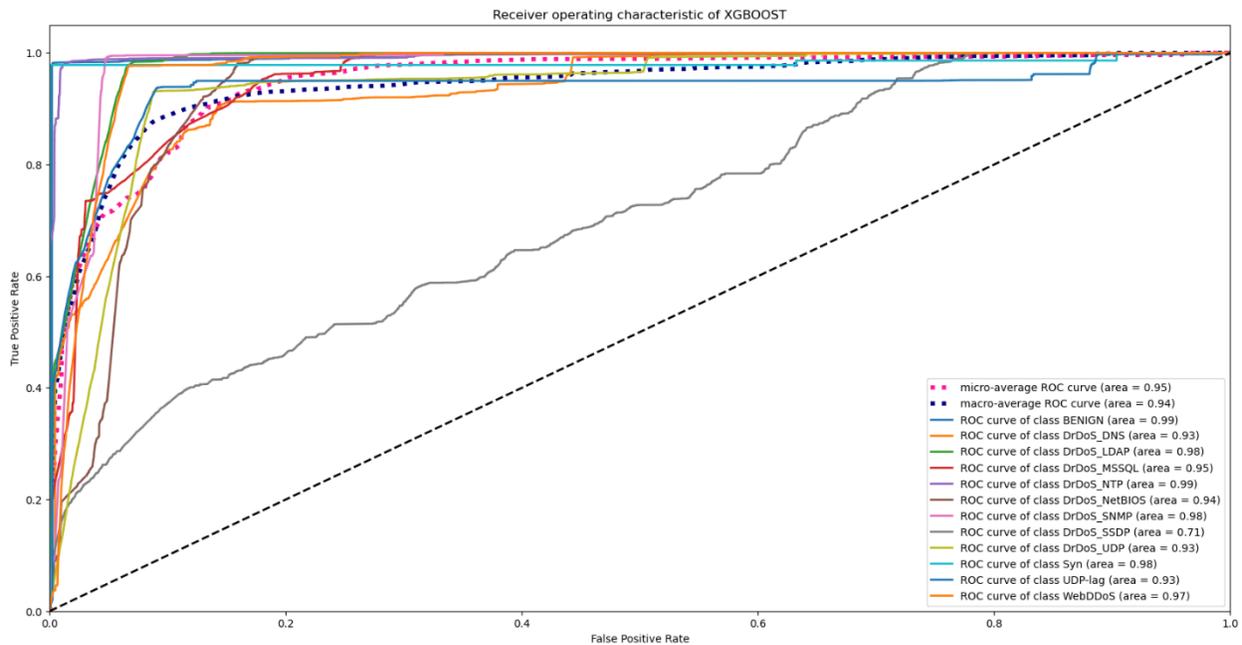

Fig. 3. ROC Curve of XGBOOST

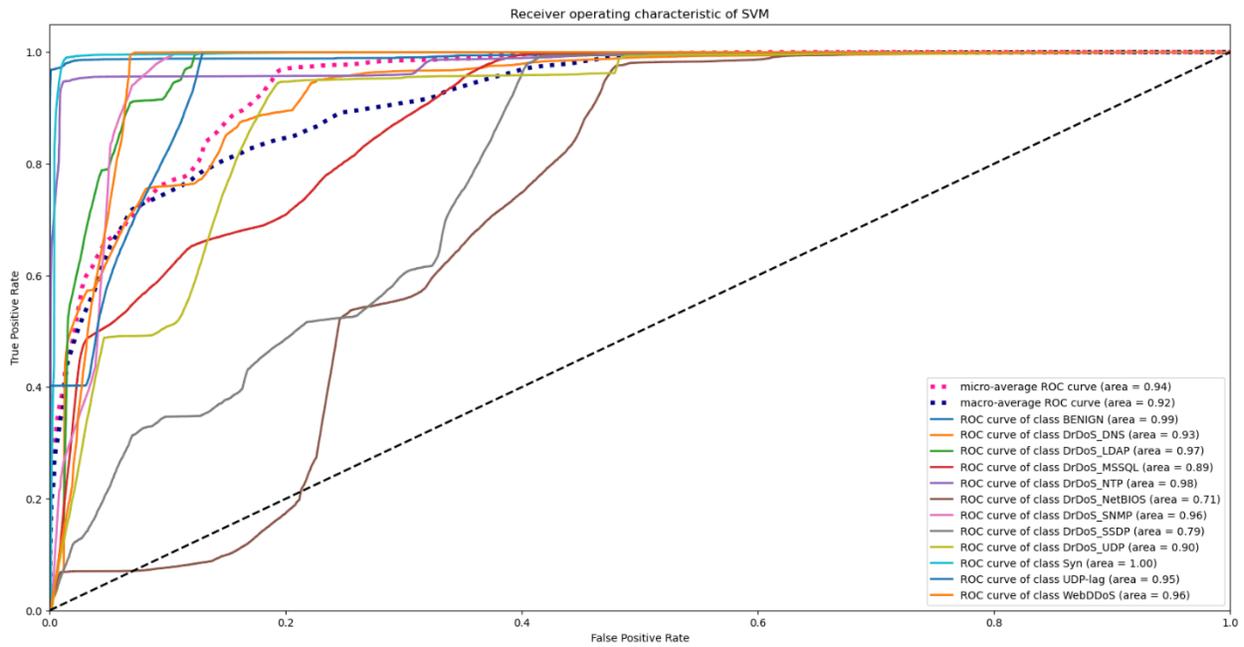

Fig. 4. ROC Curve of SVM

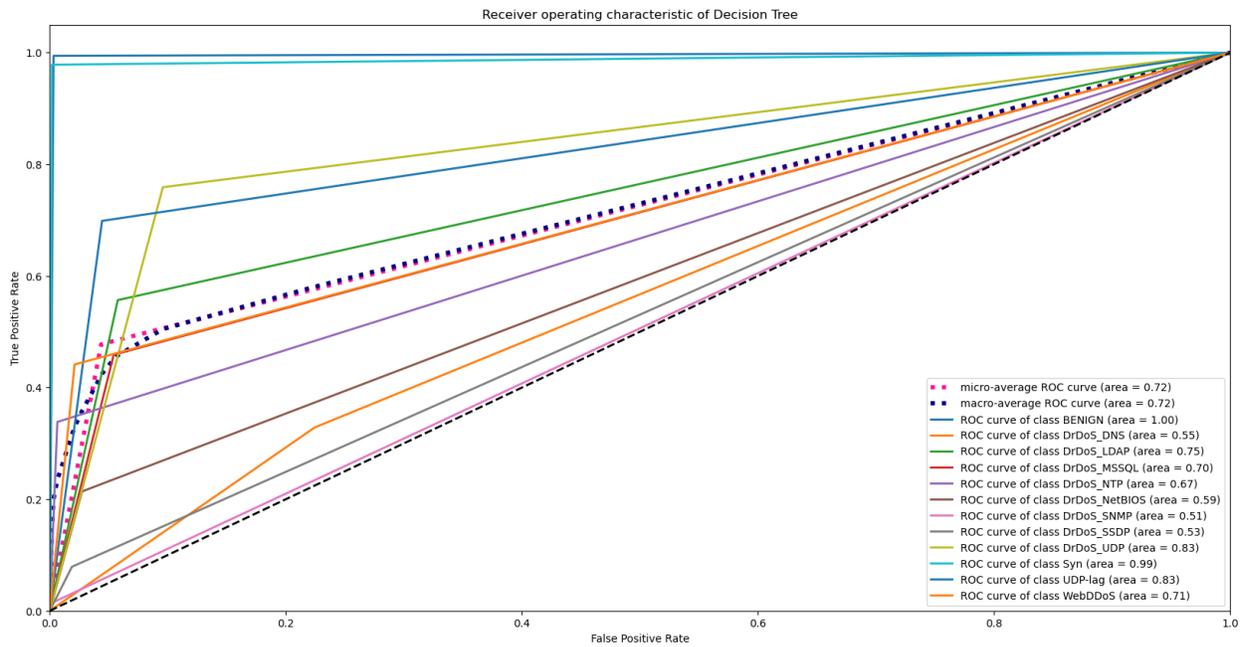

Fig. 5. ROC Curve of Decision Tree

## VI Conclusion and Future Work

This paper sought to assess the efficacy of machine learning techniques in identifying Distributed Denial of Service attacks in software-defined networking. Four algorithms, including Random Forest, Decision Tree, Support Vector

Machine, and XGBoost, were employed and evaluated utilizing the CICDDoS2019 dataset.in additional to drop some most used features as [Timestamp]. The results, evaluated through accuracy, recall, precision, and F1 score metrics, revealed that the Random Forest algorithm achieved the highest accuracy at 68.9%. The study determined that the use of ML-based detection is a more accurate and efficient approach than other methods, making it a valuable tool for actual DDoS attack identification. Despite the computational requirements of non-parametric algorithms, the results indicate their potential for providing insightful analysis of complex relationships between inputs and outputs in DDoS attack detection in SDN. Future research could focus on optimizing the computational efficiency of non-parametric algorithms, enabling their application in real-time, large-scale scenarios.

References

[1] P. Kamboj, M. C. Trivedi, V. K. Yadav, and V. K. Singh, "Detection techniques of DDoS attacks: A survey," in *2017 4th IEEE Uttar Pradesh Section International Conference on Electrical, Computer and Electronics (UPCON)*, 2017, pp. 675–679.

[2] A. Banitalebi Dehkordi, M. Soltanaghaei, and F. Z. Boroujeni, "The DDoS attacks detection through machine learning and statistical methods in SDN," *J Supercomput*, vol. 77, pp. 2383–2415, 2021.

[3] L. Wang and Y. Liu, "A DDoS attack detection method based on information entropy and deep learning in SDN," in *2020 IEEE 4th Information Technology, Networking, Electronic and Automation Control Conference (ITNEC)*, 2020, vol. 1, pp. 1084–1088.

[4] O. Rahman, M. A. G. Quraishi, and C.-H. Lung, "DDoS attacks detection and mitigation in SDN using machine learning," in *2019 IEEE world congress on services (SERVICES)*, 2019, vol. 2642, pp. 184–189.

[5] T. Abhiroop, S. Babu, and B. S. Manoj, "A machine learning approach for detecting DoS attacks in SDN switches," in *2018 Twenty Fourth National Conference on Communications (NCC)*, 2018, pp. 1–6.

[6] M. A. Aladaileh, M. Anbar, I. H. Hasbullah, Y.-W. Chong, and Y. K. Sanjalawe, "Detection techniques of distributed denial of service attacks on software-defined networking controller–a review," *IEEE Access*, vol. 8, pp. 143985–143995, 2020.

[7] D. T. Le, M. H. Dao, and Q. L. T. Nguyen, "Comparison of machine learning algorithms for DDoS attack detection in SDN," *Информационно-управляющие системы*, no. 3 (106), pp. 59–70, 2020.

[8] H. Liu and B. Lang, "Machine learning and deep learning methods for intrusion detection systems: A survey," *applied sciences*, vol. 9, no. 20, p. 4396, 2019.

[9] I. Sharafaldin, A. H. Lashkari, S. Hakak, and A. A. Ghorbani, "Developing realistic distributed denial of service (DDoS) attack dataset and taxonomy," in *2019 International Carnahan Conference on Security Technology (ICCST)*, 2019, pp. 1–8.

[10] J. Halladay *et al.*, "Detection and characterization of DDoS attacks using time-based features," *IEEE Access*, vol. 10, pp. 49794–49807, 2022.

[11] A. J. Altamemi, A. Abdulhassan, and N. T. Obeis, "DDoS attack detection in software defined networking controller using machine learning techniques," *Bulletin of Electrical Engineering and Informatics*, vol. 11, no. 5, pp. 2836–2844, 2022.

[12] F. D. S. Sumadi and C. S. K. Aditya, "Comparative Analysis of DDoS Detection Techniques Based on Machine Learning in OpenFlow Network," in *2020 3rd International Seminar on Research of Information Technology and Intelligent Systems (ISRITI)*, 2020, pp. 152–157.

[13] I. Ramadhan, P. Sukarno, and M. A. Nugroho, "Comparative analysis of K-nearest neighbor and decision tree in detecting distributed denial of service," in *2020 8th International Conference on Information and Communication Technology (ICoICT)*, 2020, pp. 1–4.

[14] K. M. Sudar, M. Beulah, P. Deepalakshmi, P. Nagaraj, and P. Chinnasamy, "Detection of Distributed Denial of Service Attacks in SDN using Machine learning techniques," in *2021 international conference on Computer Communication and Informatics (ICCCI)*, 2021, pp. 1–5.

[15] D. Firdaus, R. Munadi, and Y. Purwanto, "DDoS Attack Detection in Software Defined Network using Ensemble K-means++ and Random Forest," in *2020 3rd International Seminar on Research of Information Technology and Intelligent Systems (ISRITI)*, 2020, pp. 164–169.

[16] S. Lakshminarasimman, S. Ruswin, and K. Sundarakantham, "Detecting DDoS attacks using decision tree algorithm," in *2017 Fourth International Conference on Signal Processing, Communication and Networking (ICSCN)*, 2017, pp. 1–6.

[17] J. A. Perez-Diaz, I. A. Valdovinos, K.-K. R. Choo, and D. Zhu, "A flexible SDN-based architecture for identifying and mitigating low-rate DDoS attacks using machine learning," *IEEE Access*, vol. 8, pp. 155859–155872, 2020.

[18] A. Makuvaza, D. S. Jat, and A. M. Gamundani, "Deep neural network (DNN) solution for real-time detection of distributed denial of service (DDoS) attacks in software defined networks (SDNs)," *SN Comput Sci*, vol. 2, pp. 1–10, 2021.

[19] S. Suthaharan and S. Suthaharan, "Support vector machine," *Machine learning models and algorithms for big data classification: thinking with examples for effective learning*, pp. 207–235, 2016.

[20] Y. Liu, Y. Wang, and J. Zhang, "New machine learning algorithm: Random forest," in *Information Computing and Applications: Third International Conference, ICICA 2012, Chengde, China, September 14-16, 2012. Proceedings 3*, 2012, pp. 246–252.

[21] A. Navada, A. N. Ansari, S. Patil, and B. A. Sonkamble, "Overview of use of decision tree algorithms in machine learning," in *2011 IEEE control and system graduate research colloquium*, 2011, pp. 37–42.

[22] T. Chen and C. Guestrin, "Xgboost: A scalable tree boosting system," in *Proceedings of the 22nd acm sigkdd international conference on knowledge discovery and data mining*, 2016, pp. 785–794.

[23] L. Wasserman, *All of nonparametric statistics*. Springer Science & Business Media, 2006.

[24] P. Bühlmann and S. van de Geer, *Statistics for high-dimensional data: methods, theory and applications*. Springer Science & Business Media, 2011.

[25] M. Sokolova and G. Lapalme, "A systematic analysis of performance measures for classification tasks," *Inf Process Manag*, vol. 45, no. 4, pp. 427–437, 2009.